\documentclass[runningheads]{llncs}
\usepackage{graphicx}
\usepackage{amsmath,amssymb} 
\usepackage{color}
\usepackage[width=122mm,left=12mm,paperwidth=146mm,height=193mm,top=12mm,paperheight=217mm]{geometry}
\begin{document}
\pagestyle{headings}
\mainmatter  

\title{Learning Scene Flow With Skeleton Guidance For 3D Action Recognition} 

\author{Vasileios Magoulianitis, Athanasios Psaltis}
\institute{Centre for Research and Technology Hellas, Thessaloniki, Greece}

\maketitle

\begin{abstract}

Among the existing modalities for 3D action recognition, 3D flow has been poorly examined, although conveying rich motion information cues for human actions. Presumably, its susceptibility to noise renders it intractable, thus challenging the learning process within deep models. This work demonstrates the use of 3D flow sequence by a deep spatiotemporal model and further proposes an incremental two-level spatial attention mechanism, guided from skeleton domain, for emphasizing on features close to the body joint areas and according to their informativeness. Towards this end, an extended deep skeleton model is also introduced to learn the most discriminant action motion dynamics, so as to estimate an informativeness score for each joint. Subsequently, a late fusion scheme is adopted between the two models for learning the high level cross-modal correlations. Experimental results on the currently largest and most challenging dataset NTU RGB+D, demonstrate the effectiveness of the proposed approach, achieving state-of-the-art results.           

\end{abstract}

\section{Introduction}

The ability of computer systems to recognize human actions is of great importance, due to the broad range of its potential applications and undoubtedly has drawn much attention from research community. However, inherent challenges such as large intra-class variations or seemingly similar actions can challenge the system, especially considering possible viewpoint variations \cite{aggarwal2011human}. 

The ubiquitous accurate depth sensors such as Microsoft Kinect\textsuperscript{TM} have given rise to 3D action recognition, where depth and 3D skeleton trajectories are available apart from the appearance content. Such modalities provide robustness to illumination and appearance variations, while enriching the feature space with depth information that can be conducive for discerning ambiguous actions \cite{aggarwal2014human}. The great majority of literature approaches that have been proposed so far  \cite{Du_2015_CVPR,Vemulapalli_2014_CVPR,Trust_Gates,zhang2017geometric,huang2017deep,wang2012mining,soo2017interpretable,ke2017new,liu2017global,lee2017ensemble,li2017adaptive}
focus on using skeleton tracking data and exhibit robust performance in various datasets, while other works \cite{rahmani2016histogram,rahmani20163d,xia2013spatio,li2010action,cheng2012human} opt for using depth information or \cite{shahroudy2017deep,shahroudy2016multimodal,hu2015jointly,yu2016structure,zhao2017two,rahmani2017learning} combine more modalities to enhance the feature representation. 

Despite the remarkable performance improvements achieved by the inclusion of the optical flow in 2D action recognition task \cite{simonyan2014two,zhang2016real,feichtenhofer2016convolutional,bilen2017action}, the respective motion information in 3D space has been poorly examined. With the advent of real-time 3D flow (scene flow) estimation algorithms \cite{jaimez2015primal}, this signal can become quite attractive, since it comprises the actual motion information --by considering RGB and depth modalities-- not only for the body motion but also for the objects that human interacts with, an advantage which is missing from skeleton data. Hence, this can be detrimental for predicting fine-grained actions (e.g. ``drinking" and ``eating snacks") \cite{rahmani2017learning}, where object shape, motion and possibly affordances can help disentangling the action class \cite{thermos2017deep,koppula2013learning}. This work utilizes scene flow and represents it using the RGB colorspace as in \cite{wang2017scene}, where each motion plane is mapped on to a basic color channel. In doing so, it is favored the training process of a deep 3D spatio-temporal model (C3D) \cite{tran2015learning} that have been pre-trained using RGB data. 

Nevertheless, optical flow is admittedly a quite noisy signal \cite{eitel2015multimodal,sevilla2017integration,spies2001accurate}, posing noticeable challenges to deep learning models. Moreover, from the whole scene flow perhaps only few areas may contain really discriminative information while other parts can be misleading for action recognition. Furthermore, many studies show that not all body parts convey the same informativeness degree for this task \cite{wang2012mining,liu2017global,ofli2014sequence,raptis2012discovering,li2017action}. For instance, considering the action ``salute", from entire scene the most conducive information exists in the one hand, while the other body parts are less informative. Given that human visual cognitive system emphasizes differently on the human body parts to understand what action is performed \cite{rensink2000dynamic}, attention mechanisms have been extensively employed in numerous computer vision problems, exhibiting promising results so far \cite{cao2015look,li2018videolstm,song2017end,jaderberg2015spatial,wang2017residual,denil2012learning,NIPS2017_7181}. We propose a novel and effective mechanism to incrementally induce spatial attention in two-levels in the scene flow learning process, by using knowledge from skeleton domain to guide bottom-up the feedforward process. Therefore, this work examines the action recognition problem mainly from the scene flow perspective and especially how it can benefit from an attention mechanism which is instructed by another deep skeleton model.   

 Intuitively, the most informative parts for action recognition generally lie around joint positions as also reported by Cao et al. \cite{cao2018body} and therefore as a first step, we boost the features which lie around $2$D joints, enabling the model to pay more attention on the body joint areas. As long as not all the body parts are equally informative, the boosting degree should depend on the informativeness of each area. Temporal Convolutional Networks (TCN) have recently been proposed for action segmentation and recognition \cite{soo2017interpretable,lea2016temporal}, achieving remarkable performance and their structure can provide more insights on the important body parts for an action. We extend the TCN architecture by incorporating a new convolutional branch, namely Body Joints Convolutional Network (BJCN), which learns the discriminative movements from the entire action and its bottom layer representations can be used to estimate an informativeness score for each joint to control the boosting degree of scene flow features. Finally, skeleton and scene flow deep representations are fused with a late strategy to realize potential performance gains by learning the cross-modal coorelations, leveraging their complementary properties \cite{luo2017graph} by jointly optimizing the two models. Hence, in this setting, the skeleton model has a twofold purpose: a) learning recognizing actions and b) guiding the scene flow learning process through the attention mechanism. All in all, the major contributions are:

\begin{itemize}

\item[\textendash] A novel two-level spatial attention strategy is proposed, which uses deep skeleton domain knowledge to guide the learning process of scene flow and thus enhance its feature representation. 
\item[\textendash] An new deep skeleton model is proposed to learn the discriminant action motion dynamics, so as to extract an informativeness score for each joint.  
\item[\textendash] The high level cross-modal correlations of scene flow and skeleton deep representations are explored under the late fusion scenario.       
\item[\textendash] We carry out extensive experiments on the currently largest and most challenging dataset to validate our premises. The C3D model with attention achieves remarkably improved performance and the fused model surpasses most of the current approaches.

\end{itemize}

\section{Related Work}


\subsection{Action Recognition}

\textbf{Deep learning:} Deep learning methods prevail action recognition, especially using 3D skeleton trajectories to model the complex motion dynamics of human joints over time. Several methods \cite{Du_2015_CVPR,Trust_Gates,zhang2017geometric,liu2017global,lee2017ensemble,li2017adaptive,Shahroudy_2016_CVPR} have been proposed using variants of Recurrent Neural Networks (RNNs) for action recognition which adapt the architecture design by efficiently exploiting the physical structure of the human body or employ gating mechanisms in hidden units that enforce the spatio-temporal learning process. 
Liu {\it et al}. \cite{Trust_Gates} propose a gating mechanism for Long Short-Term Memories (LSTMs) that updates the memory cells based on the reliability of sequential data. 
Also, Li {\it et al.} \cite{li2017action} propose an action-attending neural network able to detect salient action units by adaptively weighting skeletal joints. Recently, despite the popularity of RNNs on 3D skeleton data, novel deep residual architectures have been proposed \cite{soo2017interpretable,hou2017train} that perform 1-D temporal convolutions and apart from achieving remarkable performance provide an interpretable way for action recognition. Our work in skeleton domain is based on such architectures, to propose a way for measuring an informativeness score for each joint by learning the discriminant motion dynamics for each action, an aspect that has been poorly studied from current deep learning approaches.     
   
Besides skeleton, other works employ more modalities to construct a more complete feature representation for multimodal action recognition. 
Trying to capture efficiently the correlations between RGB and depth, Shahroudy {\it et al.} \cite{shahroudy2017deep} propose a deep auto-encoder that performs common component analysis and discovers the informative parts between the two modalities. The work of  Wang {\it et al.} \cite{wang2017cooperative} illustrates how RGB and depth can be fused with two-stream deep models by using dynamic images. Moreover, \cite{rahmani2017learning,wang2017cooperative} combine skeleton with depth data to benefit from their complementary properies. Zhao {\it et al.} \cite{zhao2017two} separately learns a RNN model on skeleton data along with the C3D trained on RGB modality and lately fuse their features. However, RGB lacks depth information which in turn can hit the overall performance of the system. Another work by Luo {\it et al.} \cite{luo2017graph} proposes a graph distillation method which employs multiple modalities, demonstrating that properly utilizing them increases the performance, further confirming their complementary nature. Luo {\it et al.} \cite{luo2017unsupervised} present an unsupervised method for learning the long term motion dynamics from RGB-D input by learning a deep autoencoder to predict the 3D motion. Finally, Wang {\it et al.} \cite{wang2017scene} propose several Scene Flow to Action Maps (SFAMs) methods to represent colorized 3D flow. Yet, SFAMs include all body flow information which can be too noisy, thus challenging the deep model to learn discriminant representations of scene flow.  


\subsection{Attention Mechanisms}

 Human visual cognition selectively attends to specific parts of the scene, putting more emphasis on certain objects or colors \cite{rensink2000dynamic}. Inspired from this inherent human ability, attention mechanisms are increasingly adopted by deep learning methods on numerous problems with great success \cite{liu2017global,cao2015look,li2018videolstm,jaderberg2015spatial,wang2017residual,NIPS2017_7181}. 
For example, Jaderberg {\it et al.} \cite{jaderberg2015spatial} propose a spatial transformer module that can be integrated among layers into a CNN, confirming the attention efficiency in the house number recognition task. Another work \cite{Rama16} proposes a mechanism that boosts certain features according to their similarity, forcing the network to learn better representations for person re-identification.   
For the task of image classification, Wang {\it et al.} \cite{wang2017residual} introduce a bottom-up top-down structure to learn a mask that incurs soft attention in the more informative areas, while Cao {\it et al.} \cite{cao2015look} attain top-down attention using feedback connections. 

In 2D action recognition, Li {\it et al.} \cite{li2018videolstm} have proposed a convolutional LSTM with attention using RGB and optical flow, while Sharma {\it et al.} \cite{sharma2015action} use a multi-layer LSTM to focus on the important regions of input image which correspond with the last convolutional layer features. Regarding 3D action recognition \cite{baradel2017human}, RGB and articulated human poses are fused, along with an attention mechanism that focuses on hands. The works of Liu {\it et al.} \cite{liu2017global} and Song {\it et al.} \cite{song2017end} fall into the same category using skeleton input and building on the LSTM architecture to perform spatio-temporal attention in 3D joints motions. Among multiple ways to perform attention in a deep network, control gates have been proposed to boost certain hidden representations within deep networks and are proved to be very efficient \cite{Trust_Gates}. In our work, the attention in the feedforward process is attained though a mask, that boosts features according to their spatial position with different magnitudes, so as to focus the scene flow learning on to the informative parts, thus alleviating its noisy effects.

\section{Proposed Pipeline}

  In this section, we first present our work in deep skeleton model (Section \ref{skeleton_model}) in order to realize an informativeness score for each joint so that can be used from the attention mechanism on scene flow learning process (Section \ref{baseline_flow}). Finally, the late fused model is presented that can be jointly optimized for action recognition (Section \ref{end_to_end_model}).     
  

\subsection{Deep Skeleton Model -- Informativeness Score} \label{skeleton_model}

\begin{figure}
\centering
\includegraphics[height=6.5cm]{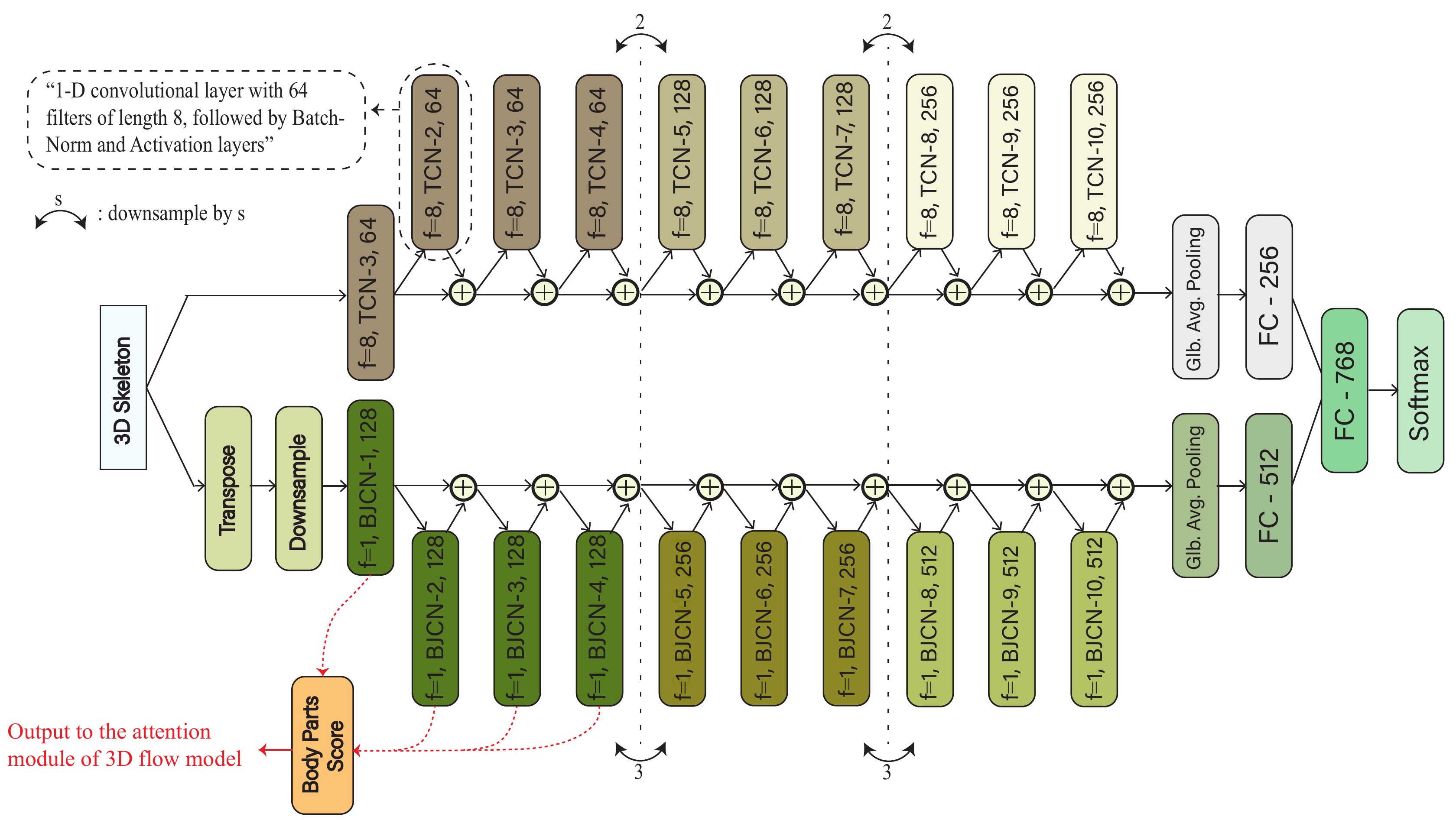}
\caption{Skeleton model Res-TCN-BJCN. The upper (TCN) and bottom (BJCN) branches are fused at the FC layers. The four bottom layers of BJCN branch are used to extract an informativeness score for each joint. Downsampling is performed by a convolutional layer (not appeared) with filter size $1$ and stride $s$ }
\label{fig:Skeleton}
\end{figure}

	 In skeleton domain, most of the works employ RNN models to learn the temporal motion patterns of body parts. Residual-TCN (Res-TCN) consists of 1D convolutional layers with residual connections and provides an efficient way for learning discriminant skeleton representations, having as key advantage that its parameters are meaningful for construing the deep model parameters.
	 Soo {\it et al.} \cite{soo2017interpretable} propose a way to interpret TCN parameters by exploiting the residual connectivity among layers, since the network is forced to learn spatio-temporal features that have direct correspondence with input feature $X_0$ as expressed from the TCN formulation:

\begin{equation}
\label{eqn:Res_TCN}
\begin{gathered}
 X^{(N)} = X^{(1)} + \sum\limits_{i=2}^{N} W^{i} * max(0, X^{(i-1)}), \\
 X^{(1)} = W_{1}X^{(0)}
\end{gathered}
\end{equation}	 
	 
 \noindent	 where $X^{(d)}$ is the feature at the $d_{th}$ convolutional layer and $X_0$ includes $M$ successively placed joint positions for $N$ frames, such that $X_0 \in \mathbb{R}^{M \times N}$.
Concretely, each convolutional layer $l$ with temporal filter size of $f_{l}$ has $N_{l}$ filters $\{ W^{i} \}_{i=1}^{N_l}$ where $W^{i} \in \mathbb{R}^{f_{l} \times N^{l-1}}$.
1-D convolutions take into account all joints and convolve the filter along time axis. So long as filters from TCN's bottom layer are of dimension $W^{i} \in \mathbb{R}^{f_{l} \times M}$, their parameters directly correspond to 3D joints, enabling a more physical interpretation.


	 Inspired from Res-TCN and having the objective to realize an informativeness score for each joint, a new model is proposed, builted on Res-TCN which apart from recognizing actions, its bottom layer features denote how much informative are the movements of each joint. Intuitively, some discriminant movements may appear in different joints and motion planes for certain actions. For instance, some joints may have steep movements (e.g. ``pushing", ``kicking"), whereas other a repetitive movement (e.g. ``salute", ``hand waving", ``shake head"). Towards this end, aiming to hierarchically learn these movements and following the deep learning paradigm, we adopt a structurally similar architecture --Eq. \ref{eqn:Res_TCN} holds also for the new network-- that shares its filter parameters across joints, mirroring the Res-TCN with another branch, namely Residual Body-Joints Convolutional Network (Res-BJCN). Each layer filters are of size $f_{l}=1$ with stride $1$ and their weights are expanded over time such that $W^{i} \in \mathbb{R}^{1 \times N}$. 
	 
	 

 
	 
	 To attain convolutions over body joints, a transpose layer is inserted after input, followed by a downsampling one along time axis to smooth the temporal sequence of joints by a factor of $5$ that experimentally proved to be optimal. Each filter of Res-BJCN actually represents a movement across the entire action and thus, downsampling facilitates filters to learn the global picture of the movement, by alleviating motion jitterings or noisy details. Following the design protocol of Res-TCN, a Batch Normalization (BN) and Rectified Linear Unit (ReLU) layers are placed after each convolutional layer. The latter one ensures that in backpropagation process the gradient flows only through the positive activations, helping bottom layers to learn better representations. Res-TCN and Res-BJCN branches are concatenated at the fully-connected (FC) layers, in order to synergistically work on action recognition. It is worth mentioning that this concatenation is imperative in order to force the Res-BJCN branch towards learning really distinctive movements, by exploiting the features of Res-TCN. The overall architecture is illustrated in Fig. \ref{fig:Skeleton}. Also, to further help the Res-BJCN branch in learning more discriminant movements, a softmax activation instead of ReLU is used after the last FC layer ($FC-512$). Finally, the number of filters is doubled when compared with TCN's branch for increasing the range of potentially learnable movements.
	
	 
	  The interesting part of this architecture is at its four bottom layers that yield features of $X \in \mathbb{R}^{M\times N_{l}}$ which represent responses for detecting at most $N_{l}$ distinct movements of every joint in 3D space. As we delve deeper in layers, each intermediate feature holds a non-linear combination of simpler movements, although after BJCN-$4$ direct correspondence is not retained due to the spatial subsampling of features. Theoretically, all features could be used to extract an informativeness score by properly upsampling them to regain correspondence, but through experiments we found that this does not significantly increase the performance and therefore we focus our examination on the features of four bottom layers. As long as filter activations denote the confidence of discriminant movements which occur in each joint, they are regarded as early signs of discriminant features because they directly contribute in higher layers through residual connections. Hence, to realize a score for each joint, filters activations are summed row-wise within a feature for the three motion planes of each join. This can be mathematically expressed as:
	  	 
\begin{equation}
\label{eqn:Score}
 score = \big \{ score_k | score_k = \sum\limits_{i=3k}^{3k+2} \sum\limits_{j=0}^{N_{l}} X_{i,j}^{(l)}, \hspace{ 2mm }  \forall k \in [0,K), \hspace{ 2mm } l \in [1,4] \big \}
\end{equation}

 \noindent where $score_k$ the informativeness score for the k-$th$ joint and $K$ the detected joints from Kinect. Finally, the vector $score \in \mathbb{R}^K$ is linearly normalized into $[0-1]$, with maximum score $score_k$ to denote that the k-$th$ joint had the more distinctive movements in 3D space against others.

 

\subsection{Baseline Flow Model} \label{baseline_flow}

	3D scene flow modality conveys instrumental properties for action recognition, accounting the real motion for all the parts of a scene. This work uses the algorithm of Jaimez {\it et al.} \cite{jaimez2015primal} for real-time flow computation using RGB and depth. A question that arises, how the dense field of 3D motion vectors can be represented as input, so that the pre-trained knowledge of powerful deep models can be better exploited, without requiring to train afresh. We opt for mapping the 3D flow vectors on to RGB colorspace, where each color channel encodes one plane of 3D motion. Specifically, scene motion vectors are linearly normalized in the interval $[0,255]$ to represent the motion magnitude with color saturations. 


 For learning the sequence of 2D colorized frames, the C3D model is employed as baseline among other options. Motivated by the analysis in \cite{tran2015learning}, we anticipate this model to learn the chromatic changes, edge orientations and how they move over time. C3D consists of $8$ convolutional layers which operate spatiotemporally on RGB sequence, interleaved by $5$ max-pooling layers. At the top of model two FC layers of output $4096$ neurons each are followed by a softmax layer with neurons equal to the number of classes. In the remainder of the paper, we follow the original notation \cite{tran2015learning} to refer in C3D layers.

\subsubsection{Attention on Boby-Parts:} \label{same_attention}

	By feeding the extracted colorized 3D flow into the C3D model, there is rich motion information for the whole scene, besides human. Noise, background motion, as well as some body-parts that are less informative are presented in the input, thus challenging the deep model on how well can adjust its parameters on the consequential parts of the action. Intuitively, motion information that lie close to human joints is more informative while the rest one can be noisy and distracting. Therefore, at the first attention level, all the body areas are equally weighted near to the $K$ detected joints. At this point, the Kinect's 2D mapped joints coordinates are used to define a circular area with radius $r$ pixels around each joint, namely ``{\em characteristic area}", since it carries a lot of properties for the joint motion. To boost characteristic areas, a binary mask is constructed which have ones in these areas and zeros elsewhere. The 3D mask is formulated as:

\begin{equation}
\label{eqn:Mask}
\begin{gathered}
M = 
\begin{cases}
m_{n,i,j}|m_{n,i,j} = 
\begin{cases}
1 \hspace{10 mm} if (i,j)  \in  \mathbb{J}_{(n)}\\
0 \hspace{10 mm} else 
\end{cases} \hspace{-3mm}
\Bigg\},
\forall n \in [0,l]
\Bigg\}
\end{cases} \hspace{-4mm}, \\
\mathbb{J}_{(n)} = 
\begin{cases}
(x,y) | (x-a)^2 + (y-b)^2 \leqslant r^2 , (a,b)  \in \mathbb{K}_{(n)}
\Big\},
n \in [0,l]
\end{cases}
\end{gathered}
\end{equation}	 

 \noindent where $\mathbb{J}_{(n)}$ is defined as the set of all characteristic areas for the n-${th}$ frame in sequence, while $\mathbb{K}_{(n)}$ is the $2$D joint positions ($\mathbb{K}_{(n)} \in  \mathbb{R}^{K \times 2}$) of that frame. Once each layer of the C3D yields a feature $Y \in \mathbb{R}^{c \times l \times h\times w}$, where $c$ the number of channels, $l$ the number of processed frames, $h$ and $w$ are the height and width of the feature, respectively, the 3D mask is finally repeated channel-wise such that $M \in \mathbb{R}^{c \times l \times h\times w}$.	

	Attention on the feedforward process is attained using a residual connection between outputted feature $Y$ from a layer and the element-wise multiplication of the feature with the constructed mask $M$, to realize a new boosted feature $F$ for feeding the next layer as follows: 
	
\begin{equation}
\label{eqn:Residual_Attention}
F = (1 + M)*Y
\end{equation}
	
\noindent The mask boosts certain features and ensures that gradient updates differently neurons that correspond to important 3D flow areas. Notably, the residual connection in the attention module serves to keep some good properties from original features in forward-pass and attenuates mask's $M$ effect on feature boosting, since good features may be out of characteristic areas (e.g. interacted objects). It is worth mentioning that this attention module can be interleaved between every pair of convolutional layers after properly downsampling the 2D joints positions to align with smaller spatially features in deeper layers. Also, to compensate the temporal max pooling size of $p = 2$, the new $2$D joint positions are realized by linearly interpolating those between two frames with step $2$, before spatial downsample (for $16$ frames, $8$ frames with new $2$D positions are realized). We place the module after all the convolutional layers of the C3D to enable feature boosting at all hierarchy levels.

\subsubsection{Weighted Attention on Body-Parts:} \label{different_attention}

\begin{figure}
\centering
\resizebox{1.0\textwidth}{!}{
\includegraphics[height=6.5cm]{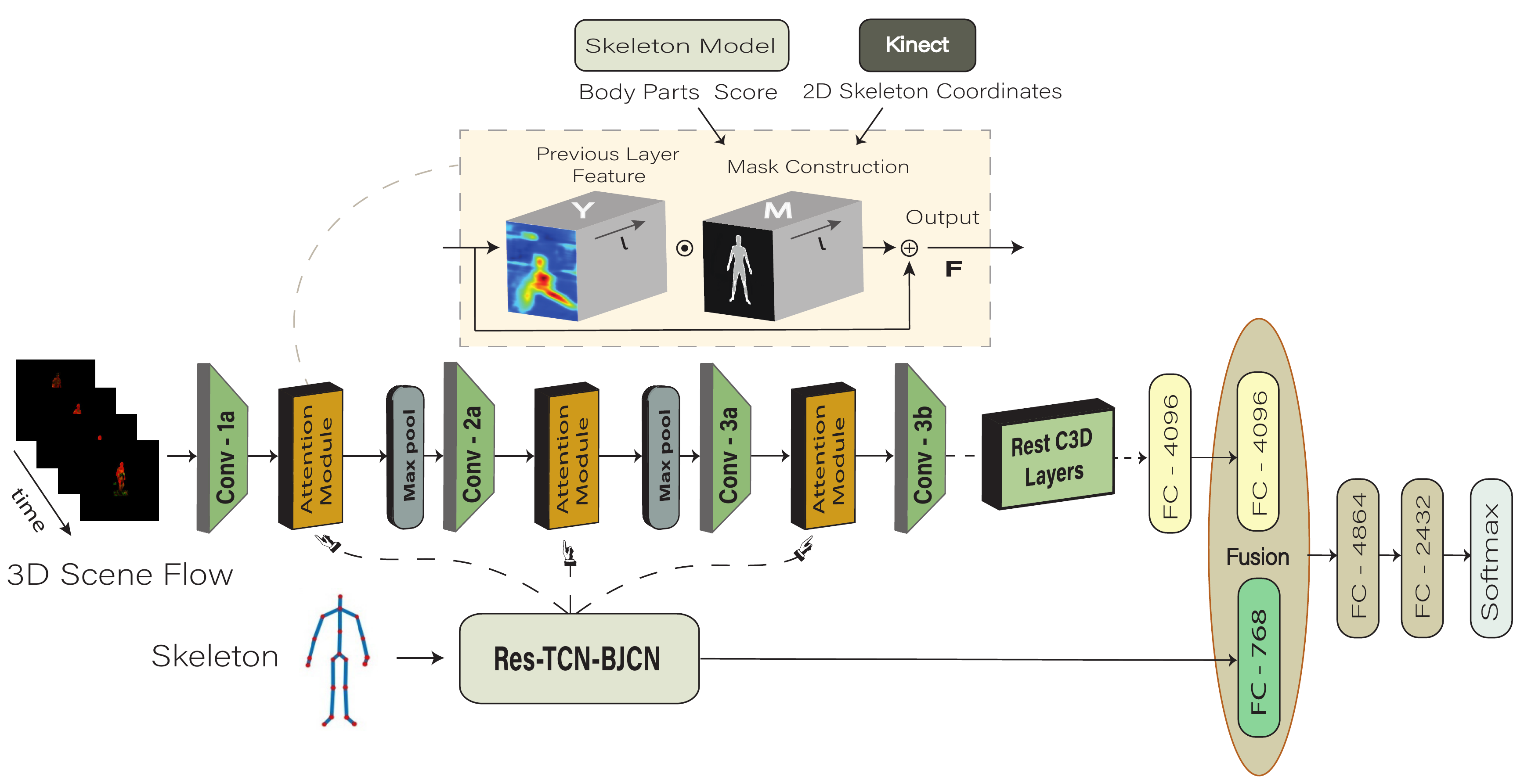}}
\caption{Fused model with skeleton and 3D flow modalities. Attention module is inserted after all convolutional layers of the C3D and receives input from Kinect and bottom layers of skeleton model. Skeleton and 3D flow modalities are lately fused by concatenating their features from FC layers}
\label{fig:end_to_end}
\end{figure}

  The first attention level $AT-1$ applied in the C3D network focuses on enhancing 3D flow feature learning close to body joint area. Since some body parts are not as much informative, the formulation of the mask should be properly adjusted in order to account accordingly the different joints. To this end, a new mask is proposed, incrementally built on the previous one, to attend on body-parts according to their informativeness, as it is imposed from deep skeleton model. Hence, the new mask will not be binary but the non-zero values in characteristic areas will be further boosted by the informativeness $score$ of the respective joint. By using the new mask, the network is instructed from skeleton deep model on which features need to be emphasised more, rather than considering that all the body-parts have the same significance for action recognition. For instance, this can be translated as ``for the action {\em kicking something} emphasize more on the features that correspond to the foot that kicks, since its motion information is more discriminant than torso". Therefore, a second spatial attention level AT-$2$ is induced in the attention mechanism incrementally on its predecessor. The residual connection within mechanism remains intact and one changes only the mask values that regulate the boosting degree of features. The new weighted 3D mask is formulated as:
 
 \begin{equation}
\label{eqn:New_Mask}
M = 
\begin{cases}
m_{n,i,j}|m_{n,i,j} = 
\begin{cases}
1 + score_k \hspace{4 mm} if (i,j)  \in  \mathbb{J}_{(n)}^k \\
0 \hspace{18 mm} else
\end{cases} \hspace{-3mm}
\Bigg\},
\forall n \in [0,l], \hspace{2mm} k \in (0,K]
\Bigg\}
\end{cases}
\end{equation}
 
\noindent where $\mathbb{J}_{(n)}^k$ the $2$D points for the k-$th$ joint's characteristic area of n-$th$ frame. It must be noted that once all the scores remain constant for the entire action, the same scores are applied to every frame of the temporal sequence, in different spatial positions though. By downsampling $2$D joints positions to fit feature spatial dimensions for deeper layers, overlaps among their characteristic areas may arise, because different joints map on to the same spatial positions in downscaled features. Therefore, the average score among the overlapped joints is retained for coordinates where two or more joints are mapped to. By averaging, we take into account the scores of all the involved joints which fall into the same receptive field, before boosting the respective features.    

	Placing the attention module between sequential CONV layers of the C3D model, as the feedforward process goes along, the features from deeper layers will be increasingly boosting, emphasizing the learning process on the more informative ones at different layers. Thus, higher layers features will be more discriminant, having repeatedly inhibited to learn parts of the signal that are less informative. Moreover, once the attention module participates in back-propagation, the mask filters the gradient in weight updating process by the factor of $(1 + M(y,\phi))$, which is mathematically expressed as follows:
	
\begin{equation}
\label{eqn:Derivative}
\frac{\partial F} {\partial \theta} = \frac{\partial ((1 + M(y,\phi)*Y(x,\theta))}{\partial \theta} = (1 + M(y,\phi))\frac{\partial Y(x,\theta)}{\partial \theta}
\end{equation}  	
	
\noindent where $x$ and $y$ is the flow and skeleton inputs parameterized by $\theta$ and $\phi$, respectively. So, the attention module regulates also how much the gradient will be boosted backwards, thus forcing it to mainly flow through the informative areas of the scene flow, considering also the max-pooling operations. 
	


\subsection{End-to-End Flow and Skeleton Deep Models} \label{end_to_end_model}

	Concerning multimodal deep architectures, features from different layers are usually combined according to the early, slow or late fusion paradigms \cite{eitel2015multimodal,rahmani2017learning}. In this way, elements from different layers are concatenated so as to learn a more complementary feature representation by learning possible correlations between the two modalities. In Section \ref{different_attention}, skeleton domain knowledge is transferred into the flow learning process, not in a traditional way via convolutional or FC layers, but is fed through the attention module for boosting commensurately the 3D flow features. Nevertheless, seeking to explore what the higher order correlations between flow and skeleton features can further provide to the final representation before classification, a late fusion approach is adopted, because its effectiveness is exhibited in various fusion problems. Therefore, the deep features of both streams (FC-$4096$ and FC-$768$) are concatenated at FC stage -softmax layers used for single-modal predictions are discarded-, in order to leverage their complementary properties. Importantly, this architecture can be jointly trained, thereby optimizing alongside the skeleton and flow models under a common objective (categorical cross-entropy minimization).
	
  By jointly optimizing both models, the learnable parameters $\phi$ (Eq. \ref{eqn:Derivative}) of the Res-TCN-BJCN bottom layers --which are involved in the informativeness score extraction-- are optimized so to improve the accuracy on action recognition using both modalities. Once these layers receive the gradient directly through the residual connections, as can be inferred from Eq. \ref{eqn:Res_TCN} when applying the chain-rule of backpropagation \cite{he2016identity}, one could argue that they are indirectly optimized also for providing better attention to the C3D model, since a high loss would affect their weights. Hence, as the training process goes along, Res-TCN-BJCN bottom layers are optimized for realizing better features for multimodal action recognition and indirectly learning also to guide the 3D flow feedforward process.	
	
	

\section{Experiments}

	The main study of this work pertains the 3D flow modality and how an attention mechanism can enhance its feature representation, especially utilizing   deep skeleton knowledge. The experimental analysis revolves around the evaluation of the baseline C3D model on the colorized flow (C3D-F) and the added value from the two attention modules, namely AT-$1$ and AT-$2$ for attention levels $1$ and $2$ respectively. We also report results for the Res-TCN-BJCN model, to evaluate the impact of Res-BJCN branch using only skeleton data, apart from feeding the attention mechanism. Finally, the lately fused model using both modalities is evaluated with two different weight update policies for drawing inferences. 

\subsection{Evaluation Dataset} \label{datasets}

\subsubsection{NTU RGB+D dataset:} This dataset \cite{Shahroudy_2016_CVPR} is currently the largest and most challenging for 3D action recognition. The large number of its samples suffice for deep learning methods to eschew overfitting issues. It contains about 4 million frames from more than $56$ thousand sequences, which typically last a few seconds, labeled in $60$ different classes and performed by $40$ subjects. The sequences are captured with three arced placed Kinect v$2$ from $80$ different viewpoints by varying the height and distance of sensors. The large viewpoint variation and intra-class distance constitutes the main challenges of the dataset. For experimentation the dataset provides $3$D skeleton sequence for each person, RGB, depth and IR modalities.    


\subsection{Implementation Details}

\subsubsection{Input preparation:}

	For the Res-TCN-BJCN the 3D joints coordinates are placed successively in a vector of size $M = 150$ (($M = 2$ skeletons) $\times$  ($K=$ 25 joints)  $\times   $ (3 dimensions)) concatenated along time axis for $N$ frames. All the raw values are linearly normalized to $[-1,1]$ for each plane of motion before fed in the deep model. The maximum length size is set to $N=300$, filling with zeros the rest frames for shorter sequences. 
	
	3D motion vectors are extracted at the Kinect's depth resolution ($512 \times 424$), so the colorized frames are of same size. To feed the C3D-F model, we center crop the 3D flow frames at size $320 \times 240$ pixels and further resize to $160 \times 120$. During training, a random cropping is performed at $112 \times 112$ to induce jittering. Each video sequence is split into $16$-frame clips with $8$ frames overlap, taking their average prediction as final result. Before splitting each sequence into clip, the flow sequence is temporally downsampled by a factor of $3$ that results in a slightly higher performance and reduces the computational burden. Therefore the C3D-F receives input clips of size $3 \times 16 \times 112 \times 112$. 
	 
\subsubsection{Training and testing:}

For both skeleton and C3D models the Keras deep learning framework \cite{chollet2015keras} with Tensorflow backend \cite{abadi2016tensorflow} was used for experimentation on two Nvidia Tesla K$40$ GPU. In Res-TCN-BJCN model the initial learning rate ($lr$) is set at $1e^{-2}$, diminishing it by a factor of $10$ when the testing loss plateaus for more than $10$ epochs. At all convolutional layers a $L1$ regularizer is applied with a weight of $1e^{-4}$ and their learnable parameters are initialized with Glorot's algorithm \cite{glorot2010understanding}. After all activation layers, we use dropout with $0.5$ rate to anticipate overfitting effects. For optimizing the network, we use Stochastic Gradient Descent (SGD) with nesterov acceleration and momentum $0.9$. The batch size is set to $128$ and the model is trained from scratch for $150$ epochs. 
	
To train the baseline C3D-F model on 3D flow data, we use its pre-trained parameters from Sports-1M dataset \cite{karpathy2014large}. Specifically, the baseline C3D-F and its extension with the first level of attention module is trained with SGD using mini-batches of $60$ clips with $lr = 3e^{-3}$. The $lr$ is divided by $5$ every $4$ epochs and the model requires $30$ epochs to converge. For jointly training the fused model, the pre-trained on NTU RGB+D skeleton model is used, since its bottom layer's parameters need to be already trained for efficiently guiding the C3D from early epochs. The $r$ hyper-parameter in attention module that controls the characteristic area span, experimentally found to be optimal when $r = \lceil 0.05 \times w \rceil$, where $w$ is the width of the input feature to the attention module.

\subsection{Results and discussion}
	
\setlength{\tabcolsep}{4.0 pt}	
 \begin{table}[!htb]
    \begin{minipage}{.45\linewidth}
      \centering
       \caption{Performance accuracy (\%) on NTU RGB+D}
      	  \label{table:Results}
		  \begin{tabular}{|c|c|c|}
			\hline
			Method  & CS & CV	\\
			\hline
			 \multicolumn{3}{|c|}{\textbf{Skeleton}} \\
			\hline
			Lie Group \cite{Vemulapalli_2014_CVPR} & $50.1 $ & $52.8 $\\
			Part-aware LSTM \cite{Shahroudy_2016_CVPR} & $62.9 $ & $70.3 $ \\
			Trust Gate \cite{Trust_Gates} & $69.2 $ & $77.7 $ \\
			STA-LSTM \cite{song2017end} & $62.9 $ & $70.3 $ \\
			GCA-LSTM \cite{liu2017global} & $73.4 $ & $81.2 $ \\
			Res-TCN \cite{soo2017interpretable} & $74.3 $ & $83.1 $ \\
			Adaptive-Tree \cite{li2017adaptive} & $74.6$ & $83.2$ \\
			MTLN \cite{ke2017new} & $\mathbf{79.6}$ &	$84.8 $ \\
			\textbf{Res-TCN-BJCN} & $76.0 $ & $\mathbf{85.1}$ \\
			\hline
			 \multicolumn{3}{|c|}{\textbf{Flow}} \\
			\hline
			SFAM \cite{wang2017scene} & $57.4 $	& $59.1 $ \\
			Atomic3DFlow \cite{luo2017unsupervised} & $66.2$ & - \\
			\textbf{C3D-F (from scratch)} & $73.5 $	& $78.8 $ \\
			\textbf{C3D-F (pre-trained)} &	$75.4 $	& $81.4 $ \\
			\textbf{C3D-F + AT-$\mathbf{1}$} & $ 78.2 $ & $84.5 $ \\
			\textbf{C3D-F + AT-$\mathbf{2}$} & $ \mathbf{80.3}$ & $\mathbf{87.3}$ \\
			\hline
			 \multicolumn{3}{|c|}{\textbf{Fused}} \\
			\hline
			FTP Dyn. Skeleton \cite{hu2015jointly} &	$60.2$	& $65.2$ \\
			SSSCA-SSLM \cite{shahroudy2017deep} &	$74.9$ &  -	\\
			STA-Hands \cite{baradel2017human} & $82.5$ & $88.6$ \\
			Graph-distillation \cite{luo2017graph} & $\mathbf{89.5}$ & -	\\  
			\textbf{Late fused model} & $84.2 $	& $\mathbf{90.3}$ \\
			\hline
		\end{tabular}
    \end{minipage}%
    \hspace{5mm}
    \begin{minipage}{.5\linewidth}
      \centering
        \caption{Comparative results on C3D-F model accuracy (\%) drawing features from different bottom layers of Res-BJCN branch to realize the informativeness vector}
        \label{table:Bottom_layers}
		\begin{tabular}{|c|c|c|}
			\hline
			Layer  & CS & CV	\\
			\hline
			BJCN-1 & $79.0 $	& $85.6$ \\
			BJCN-2 & $79.2 $	& $85.9 $ \\
			BJCN-3 & $79.8$ & $86.5$ \\
			BJCN-4 & $79.6$	& $86.8$ \\
			\hline
			Sum & $\mathbf{80.3}$	& $\mathbf{87.3}$ \\
			\hline
		\end{tabular}

		\vspace{32 mm}

      \centering
        \caption{Fused model accuracy (\%) with two gradient update policies: ($1$) excluding skeleton model from updating its weights or ($2$) jointly optimizing both models}
        \label{table:Freezing}
		\begin{tabular}{|c|c|c|}
			\hline
			Method  & CS & CV	\\
			\hline
			Freezed Skeleton ($1$) & $81.0$	& $87.8$ \\
			Joint ($2$)  & $\mathbf{84.2}$ & $\mathbf{90.3}$ \\
			\hline
			
		\end{tabular}
		
    \end{minipage} 
\end{table}	
\setlength{\tabcolsep}{1.4pt}

 NTU RGB+D dataset has two evaluation standard protocols: {\em cross-subject} where training set is formed using samples of $20$ subjects and test set with the remaining ones, and {\em cross-view} where samples from two cameras are used for training and the third one for testing. In Table \ref{table:Results}, the results in terms of precision accuracy are provided for both evaluation protocols.


To begin with, the C3D-F baseline achieves satisfactory accuracy, especially pre-trained, demonstrating the rich properties of scene flow modality, yet confined to learn optimal features due to its aforementioned limitations. To answer the question whether the proposed attention mechanisms is effective the results are convincing. The incorporation of AT-$1$ module improves by significant margin ($+3.1 \%$ for cross-view) over the baseline, demonstrating that indeed from the entire scene flow, the characteristic areas around joints are considerably informative, making background or other body areas somewhat distracting. Moving on to next level (AT-$2$), the accuracy is further improved ($+2.5 \%$ on average), validating our initial hypothesis that different emphasis should be paid on the different body-parts. Notably, the AT-$2$ mechanism yields slightly better improvements for the cross-view protocol ($+0.7 \%$), denoting that is a bit more effective on dealing with viewpoint variations. The C3D-F + AT-$2$ surpasses most current state-of-the art approaches, confirming that 3D flow is a quite informative modality but entails special treatment within learning process

On the other side, the initial objective of Res-BJCN branch meant to realize an informativeness score for each joint. As a side benefit, this amendment increases the performance of skeleton model by roughly $+2 \%$, showing that the new features enhance the deep skeleton representation. 

As earlier discussed, the four bottom layers of Res-BJCN branch can be used for realizing a score for each joint, due to their direct correspondence with input. So, we experimented by using each of them to infer which learns better representations for guiding the C3D-F. As can be seen in Table \ref{table:Bottom_layers}, deeper bottom layers $3$ and $4$ provide better features to the attention module. It implies that filters of these layers have learned a non-linear combination of simpler movements from previous layers that represent more complex ones and their responses are more reliable for guiding the 3D flow learning process. In particular, the best results are yielded by summarizing all four bottom layer features and afterwards scaling into $[0-1]$ to realize the score, showing that their combination have better properties rather than using each one separately.  
	
 The final experiment on fused model validates that heterogeneous features, though naively fused, can enrich the feature space, thereby increasing the overall performance by large margins. Other attempts for classifying the fused features (SVM and naive Bayes classifiers) proved inferior in terms of accuracy. Furthermore, to evaluate the impact of Res-TCN-BJCN in the whole pipeline, we conduct experiments on the fused model, either excluding its weights from being trained (up to FC-$768$) or jointly optimizing both models. The results in Table \ref{table:Freezing} confirm that by optimizing alongside both models, results in considerably improved accuracy, by leveraging the higher order cross-modal correlations.  
 

	    
\begin{figure}
\centering
\resizebox{1.0\textwidth}{!}{
\includegraphics[height=6.5cm]{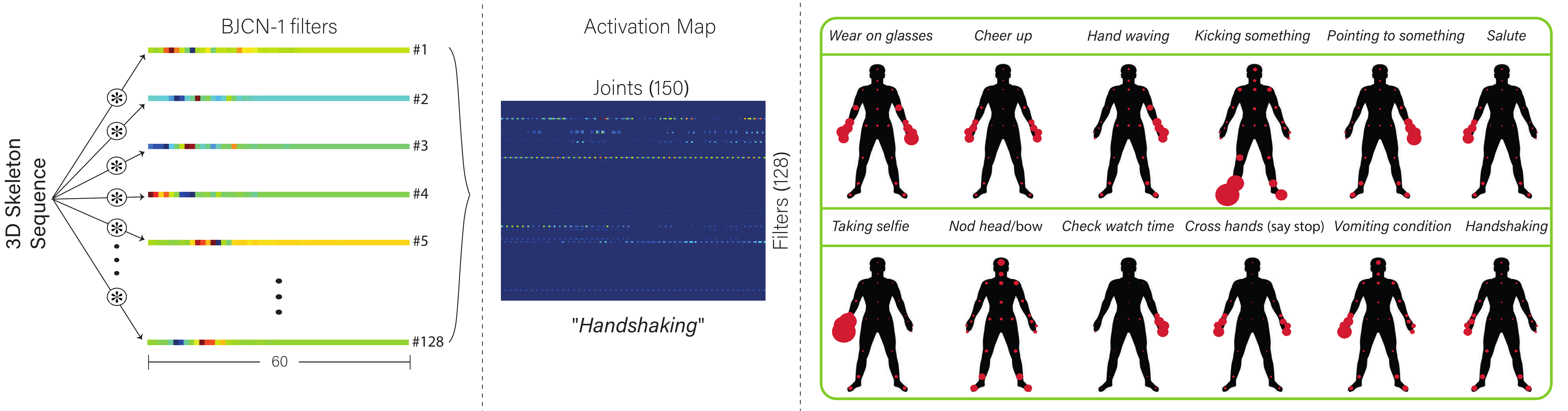}}
\caption{Visualization of the informativeness score extraction pipeline. The top-$6$ responsive filters are illustrated from BJCN-$1$ layer (left), the activation map (transposed) after convolving them on skeleton input (center) and the derived scores of body parts for some indicative actions (right). The larger the red colored circle, the higher the value of informativeness score. \textbf{Best viewed in color}}
\label{fig:Visualization}
\end{figure}

Finally, in order to provide more insights on what emphasis is given at the AT-$2$ on the different body parts, we provide visualizations for the informativeness score extraction pipeline. In Fig. \ref{fig:Visualization}, it is illustrated that bottom layer filters have learned some basic motion dynamics of joints, where only a subset of them is responsive for certain actions, thus validating the improvements incurred from the attention mechanism. We can see that discriminant movements are occurred mainly in hands or feet according to the type of action. For instance, in action {\em ``kicking something"}, the right foot which kicks or the pivoting one have more conducive information, although all body is moving, while for {\em ``salute"} and {\em ``taking selfie"} classes the right hand is more informative. 


\section{Conclusion}

In this paper, the 3D flow modality is examined for action recognition and how its representation can be enhanced using attention mechanisms guided by deep skeleton representations. Once 3D flow has spatial areas that are less informative than others, a two-level spatial attention mechanism is proposed that boosts certain features which lie around body joints and according to their informativeness for skeleton domain. For this purpose, a new skeleton model is proposed that its bottom feature representations can be used to estimate an informativeness score for each joint. Finally, the skeleton and flow deep features are fused in a later stage for jointly optimizing the whole network using both modalities and learning their higher order cross-modal correlations. The provided experimental results and visualizations validate our initial premises on the proposed attention process in 3D flow learning which achieves superior performance over other approaches. As future work, we plan to extend the attention mechanism on the temporal domain, by properly revisiting the skeleton model and also to explore more sophisticated fusion schemes.

\bibliographystyle{splncs}
\bibliography{refs}

\begin{thebibliography}{10}

\bibitem{aggarwal2011human}
Aggarwal, J.K., Ryoo, M.S.:
\newblock Human activity analysis: A review.
\newblock ACM Computing Surveys (CSUR) \textbf{43}(3) (2011) ~16

\bibitem{aggarwal2014human}
Aggarwal, J.K., Xia, L.:
\newblock Human activity recognition from 3d data: A review.
\newblock Pattern Recognition Letters \textbf{48} (2014)  70--80

\bibitem{Du_2015_CVPR}
Du, Y., Wang, W., Wang, L.:
\newblock Hierarchical recurrent neural network for skeleton based action
  recognition.
\newblock In: The IEEE Conference on Computer Vision and Pattern Recognition
  (CVPR). (June 2015)

\bibitem{Vemulapalli_2014_CVPR}
Vemulapalli, R., Arrate, F., Chellappa, R.:
\newblock Human action recognition by representing 3d skeletons as points in a
  lie group.
\newblock In: The IEEE Conference on Computer Vision and Pattern Recognition
  (CVPR). (June 2014)

\bibitem{Trust_Gates}
Liu, J., Shahroudy, A., Xu, D., Wang, G.:
\newblock Spatio-temporal lstm with trust gates for 3d human action
  recognition.
\newblock In Leibe, B., Matas, J., Sebe, N., Welling, M., eds.: Computer Vision
  -- ECCV 2016, Cham, Springer International Publishing (2016)  816--833

\bibitem{zhang2017geometric}
Zhang, S., Liu, X., Xiao, J.:
\newblock On geometric features for skeleton-based action recognition using
  multilayer lstm networks.
\newblock In: Applications of Computer Vision (WACV), 2017 IEEE Winter
  Conference on, IEEE (2017)  148--157

\bibitem{huang2017deep}
Huang, Z., Wan, C., Probst, T., Van~Gool, L.:
\newblock Deep learning on lie groups for skeleton-based action recognition.
\newblock In: Proceedings of the 2017 IEEE Conference on Computer Vision and
  Pattern Recognition (CVPR), IEEE computer Society (2017)  6099--6108

\bibitem{wang2012mining}
Wang, J., Liu, Z., Wu, Y., Yuan, J.:
\newblock Mining actionlet ensemble for action recognition with depth cameras.
\newblock In: Computer Vision and Pattern Recognition (CVPR), 2012 IEEE
  Conference on, IEEE (2012)  1290--1297

\bibitem{soo2017interpretable}
Soo~Kim, T., Reiter, A.:
\newblock Interpretable 3d human action analysis with temporal convolutional
  networks.
\newblock In: Proceedings of the IEEE Conference on Computer Vision and Pattern
  Recognition Workshops. (2017)  20--28

\bibitem{ke2017new}
Ke, Q., Bennamoun, M., An, S., Sohel, F., Boussaid, F.:
\newblock A new representation of skeleton sequences for 3d action recognition.
\newblock arXiv preprint arXiv:1703.03492 (2017)

\bibitem{liu2017global}
Liu, J., Wang, G., Hu, P., Duan, L.Y., Kot, A.C.:
\newblock Global context-aware attention lstm networks for 3d action
  recognition.
\newblock In: CVPR. (2017)

\bibitem{lee2017ensemble}
Lee, I., Kim, D., Kang, S., Lee, S.:
\newblock Ensemble deep learning for skeleton-based action recognition using
  temporal sliding lstm networks.
\newblock In: 2017 IEEE International Conference on Computer Vision (ICCV),
  IEEE (2017)  1012--1020

\bibitem{li2017adaptive}
Li, W., Wen, L., Chang, M.C., Lim, S.N., Lyu, S.:
\newblock Adaptive rnn tree for large-scale human action recognition.
\newblock In: Proceedings of the IEEE Conference on Computer Vision and Pattern
  Recognition. (2017)  1444--1452

\bibitem{rahmani2016histogram}
Rahmani, H., Mahmood, A., Huynh, D., Mian, A.:
\newblock Histogram of oriented principal components for cross-view action
  recognition.
\newblock IEEE transactions on pattern analysis and machine intelligence
  \textbf{38}(12) (2016)  2430--2443

\bibitem{rahmani20163d}
Rahmani, H., Mian, A.:
\newblock 3d action recognition from novel viewpoints.
\newblock In: Proceedings of the IEEE Conference on Computer Vision and Pattern
  Recognition. (2016)  1506--1515

\bibitem{xia2013spatio}
Xia, L., Aggarwal, J.:
\newblock Spatio-temporal depth cuboid similarity feature for activity
  recognition using depth camera.
\newblock In: Proceedings of the IEEE Conference on Computer Vision and Pattern
  Recognition. (2013)  2834--2841

\bibitem{li2010action}
Li, W., Zhang, Z., Liu, Z.:
\newblock Action recognition based on a bag of 3d points.
\newblock In: Computer Vision and Pattern Recognition Workshops (CVPRW), 2010
  IEEE Computer Society Conference on, IEEE (2010)  9--14

\bibitem{cheng2012human}
Cheng, Z., Qin, L., Ye, Y., Huang, Q., Tian, Q.:
\newblock Human daily action analysis with multi-view and color-depth data.
\newblock In: Computer Vision--ECCV 2012. Workshops and Demonstrations,
  Springer (2012)  52--61

\bibitem{shahroudy2017deep}
Shahroudy, A., Ng, T.T., Gong, Y., Wang, G.:
\newblock Deep multimodal feature analysis for action recognition in rgb+ d
  videos.
\newblock IEEE Transactions on Pattern Analysis and Machine Intelligence (2017)

\bibitem{shahroudy2016multimodal}
Shahroudy, A., Ng, T.T., Yang, Q., Wang, G.:
\newblock Multimodal multipart learning for action recognition in depth videos.
\newblock IEEE transactions on pattern analysis and machine intelligence
  \textbf{38}(10) (2016)  2123--2129

\bibitem{hu2015jointly}
Hu, J.F., Zheng, W.S., Lai, J., Zhang, J.:
\newblock Jointly learning heterogeneous features for rgb-d activity
  recognition.
\newblock In: Proceedings of the IEEE conference on computer vision and pattern
  recognition. (2015)  5344--5352

\bibitem{yu2016structure}
Yu, M., Liu, L., Shao, L.:
\newblock Structure-preserving binary representations for rgb-d action
  recognition.
\newblock IEEE transactions on pattern analysis and machine intelligence
  \textbf{38}(8) (2016)  1651--1664

\bibitem{zhao2017two}
Zhao, R., Ali, H., van~der Smagt, P.:
\newblock Two-stream rnn/cnn for action recognition in 3d videos.
\newblock Intelligent Robots and Systems (IROS), 2015 IEEE/RSJ International
  Conference on (2017)

\bibitem{rahmani2017learning}
Rahmani, H., Bennamoun, M.:
\newblock Learning action recognition model from depth and skeleton videos.
\newblock In: Proceedings of the IEEE Conference on Computer Vision and Pattern
  Recognition. (2017)  5832--5841

\bibitem{simonyan2014two}
Simonyan, K., Zisserman, A.:
\newblock Two-stream convolutional networks for action recognition in videos.
\newblock In: Advances in neural information processing systems. (2014)
  568--576

\bibitem{zhang2016real}
Zhang, B., Wang, L., Wang, Z., Qiao, Y., Wang, H.:
\newblock Real-time action recognition with enhanced motion vector cnns.
\newblock In: Proceedings of the IEEE Conference on Computer Vision and Pattern
  Recognition. (2016)  2718--2726

\bibitem{feichtenhofer2016convolutional}
Feichtenhofer, C., Pinz, A., Zisserman, A.:
\newblock Convolutional two-stream network fusion for video action recognition.
\newblock In: Proceedings of the IEEE Conference on Computer Vision and Pattern
  Recognition. (2016)  1933--1941

\bibitem{bilen2017action}
Bilen, H., Fernando, B., Gavves, E., Vedaldi, A.:
\newblock Action recognition with dynamic image networks.
\newblock IEEE Transactions on Pattern Analysis and Machine Intelligence (2017)

\bibitem{jaimez2015primal}
Jaimez, M., Souiai, M., Gonzalez-Jimenez, J., Cremers, D.:
\newblock A primal-dual framework for real-time dense rgb-d scene flow.
\newblock In: Robotics and Automation (ICRA), 2015 IEEE International
  Conference on, IEEE (2015)  98--104

\bibitem{thermos2017deep}
Thermos, S., Papadopoulos, G.T., Daras, P., Potamianos, G.:
\newblock Deep affordance-grounded sensorimotor object recognition.
\newblock In: Proceedings of the IEEE Conference on Computer Vision and Pattern
  Recognition. (2017)

\bibitem{koppula2013learning}
Koppula, H.S., Gupta, R., Saxena, A.:
\newblock Learning human activities and object affordances from rgb-d videos.
\newblock The International Journal of Robotics Research \textbf{32}(8) (2013)
  951--970

\bibitem{wang2017scene}
Wang, P., Li, W., Gao, Z., Zhang, Y., Tang, C., Ogunbona, P.:
\newblock Scene flow to action map: A new representation for rgb-d based action
  recognition with convolutional neural networks.
\newblock In: Proceedings of the IEEE Conference on Computer Vision and Pattern
  Recognition. (2017)

\bibitem{tran2015learning}
Tran, D., Bourdev, L., Fergus, R., Torresani, L., Paluri, M.:
\newblock Learning spatiotemporal features with 3d convolutional networks.
\newblock In: Proceedings of the IEEE international conference on computer
  vision. (2015)  4489--4497

\bibitem{eitel2015multimodal}
Eitel, A., Springenberg, J.T., Spinello, L., Riedmiller, M., Burgard, W.:
\newblock Multimodal deep learning for robust rgb-d object recognition.
\newblock In: Intelligent Robots and Systems (IROS), 2015 IEEE/RSJ
  International Conference on, IEEE (2015)  681--687

\bibitem{sevilla2017integration}
Sevilla-Lara, L., Liao, Y., Guney, F., Jampani, V., Geiger, A., Black, M.J.:
\newblock On the integration of optical flow and action recognition.
\newblock arXiv preprint arXiv:1712.08416 (2017)

\bibitem{spies2001accurate}
Spies, H., Scharr, H.:
\newblock Accurate optical flow in noisy image sequences.
\newblock In: Computer Vision, 2001. ICCV 2001. Proceedings. Eighth IEEE
  International Conference on. Volume~1., IEEE (2001)  587--592

\bibitem{ofli2014sequence}
Ofli, F., Chaudhry, R., Kurillo, G., Vidal, R., Bajcsy, R.:
\newblock Sequence of the most informative joints (smij): A new representation
  for human skeletal action recognition.
\newblock Journal of Visual Communication and Image Representation
  \textbf{25}(1) (2014)  24--38

\bibitem{raptis2012discovering}
Raptis, M., Kokkinos, I., Soatto, S.:
\newblock Discovering discriminative action parts from mid-level video
  representations.
\newblock In: Computer Vision and Pattern Recognition (CVPR), 2012 IEEE
  Conference on, IEEE (2012)  1242--1249

\bibitem{li2017action}
Li, C., Cui, Z., Zheng, W., Xu, C., Ji, R., Yang, J.:
\newblock Action-attending graphic neural network.
\newblock arXiv preprint arXiv:1711.06427 (2017)

\bibitem{rensink2000dynamic}
Rensink, R.A.:
\newblock The dynamic representation of scenes.
\newblock Visual cognition \textbf{7}(1-3) (2000)  17--42

\bibitem{cao2015look}
Cao, C., Liu, X., Yang, Y., Yu, Y., Wang, J., Wang, Z., Huang, Y., Wang, L.,
  Huang, C., Xu, W.,  et~al.:
\newblock Look and think twice: Capturing top-down visual attention with
  feedback convolutional neural networks.
\newblock In: Proceedings of the IEEE International Conference on Computer
  Vision. (2015)  2956--2964

\bibitem{li2018videolstm}
Li, Z., Gavrilyuk, K., Gavves, E., Jain, M., Snoek, C.G.:
\newblock Videolstm convolves, attends and flows for action recognition.
\newblock Computer Vision and Image Understanding \textbf{166} (2018)  41--50

\bibitem{song2017end}
Song, S., Lan, C., Xing, J., Zeng, W., Liu, J.:
\newblock An end-to-end spatio-temporal attention model for human action
  recognition from skeleton data.
\newblock In: AAAI. (2017)  4263--4270

\bibitem{jaderberg2015spatial}
Jaderberg, M., Simonyan, K., Zisserman, A.,  et~al.:
\newblock Spatial transformer networks.
\newblock In: Advances in Neural Information Processing Systems. (2015)
  2017--2025

\bibitem{wang2017residual}
Wang, F., Jiang, M., Qian, C., Yang, S., Li, C., Zhang, H., Wang, X., Tang, X.:
\newblock Residual attention network for image classification.
\newblock (2017)

\bibitem{denil2012learning}
Denil, M., Bazzani, L., Larochelle, H., de~Freitas, N.:
\newblock Learning where to attend with deep architectures for image tracking.
\newblock Neural computation \textbf{24}(8) (2012)  2151--2184

\bibitem{NIPS2017_7181}
Vaswani, A., Shazeer, N., Parmar, N., Uszkoreit, J., Jones, L., Gomez, A.N.,
  Kaiser, L.u., Polosukhin, I.:
\newblock Attention is all you need.
\newblock In Guyon, I., Luxburg, U.V., Bengio, S., Wallach, H., Fergus, R.,
  Vishwanathan, S., Garnett, R., eds.: Advances in Neural Information
  Processing Systems 30.
\newblock Curran Associates, Inc. (2017)  6000--6010

\bibitem{cao2018body}
Cao, C., Zhang, Y., Zhang, C., Lu, H.:
\newblock Body joint guided 3-d deep convolutional descriptors for action
  recognition.
\newblock IEEE transactions on cybernetics \textbf{48}(3) (2018)  1095--1108

\bibitem{lea2016temporal}
Lea, C., Flynn, M.D., Vidal, R., Reiter, A., Hager, G.D.:
\newblock Temporal convolutional networks for action segmentation and
  detection.
\newblock arXiv preprint arXiv:1611.05267 (2016)

\bibitem{luo2017graph}
Luo, Z., Jiang, L., Hsieh, J.T., Niebles, J.C., Fei-Fei, L.:
\newblock Graph distillation for action detection with privileged information.
\newblock arXiv preprint arXiv:1712.00108 (2017)

\bibitem{Shahroudy_2016_CVPR}
Shahroudy, A., Liu, J., Ng, T.T., Wang, G.:
\newblock Ntu rgb+d: A large scale dataset for 3d human activity analysis.
\newblock In: The IEEE Conference on Computer Vision and Pattern Recognition
  (CVPR). (June 2016)

\bibitem{hou2017train}
Hou, J., Kim, T.S., Reiter, A.:
\newblock Train, diagnose and fix: Interpretable approach for fine-grained
  action recognition.
\newblock arXiv preprint arXiv:1711.08502 (2017)

\bibitem{wang2017cooperative}
Wang, P., Li, W., Wan, J., Ogunbona, P., Liu, X.:
\newblock Cooperative training of deep aggregation networks for rgb-d action
  recognition.
\newblock arXiv preprint arXiv:1801.01080 (2017)

\bibitem{luo2017unsupervised}
Luo, Z., Peng, B., Huang, D.A., Alahi, A., Fei-Fei, L.:
\newblock Unsupervised learning of long-term motion dynamics for videos.
\newblock arXiv preprint arXiv:1701.01821 (2017)

\bibitem{Rama16}
Varior, R.R., Haloi, M., Wang, G.:
\newblock Gated siamese convolutional neural network architecture for human
  re-identification.
\newblock In Leibe, B., Matas, J., Sebe, N., Welling, M., eds.: Computer Vision
  -- ECCV 2016, Cham, Springer International Publishing (2016)  791--808

\bibitem{sharma2015action}
Sharma, S., Kiros, R., Salakhutdinov, R.:
\newblock Action recognition using visual attention.
\newblock (2016)

\bibitem{baradel2017human}
Baradel, F., Wolf, C., Mille, J.:
\newblock Human action recognition: Pose-based attention draws focus to hands.
\newblock In: ICCV Workshop on Hands in Action. (2017)

\bibitem{he2016identity}
He, K., Zhang, X., Ren, S., Sun, J.:
\newblock Identity mappings in deep residual networks.
\newblock In: European Conference on Computer Vision, Springer (2016)  630--645

\bibitem{chollet2015keras}
Chollet, F.,  et~al.:
\newblock Keras (2015)

\bibitem{abadi2016tensorflow}
Abadi, M., Barham, P., Chen, J., Chen, Z., Davis, A., Dean, J., Devin, M.,
  Ghemawat, S., Irving, G., Isard, M.,  et~al.:
\newblock Tensorflow: A system for large-scale machine learning.
\newblock In: OSDI. Volume~16. (2016)  265--283

\bibitem{glorot2010understanding}
Glorot, X., Bengio, Y.:
\newblock Understanding the difficulty of training deep feedforward neural
  networks.
\newblock In: Proceedings of the Thirteenth International Conference on
  Artificial Intelligence and Statistics. (2010)  249--256

\bibitem{karpathy2014large}
Karpathy, A., Toderici, G., Shetty, S., Leung, T., Sukthankar, R., Fei-Fei, L.:
\newblock Large-scale video classification with convolutional neural networks.
\newblock In: Proceedings of the IEEE conference on Computer Vision and Pattern
  Recognition. (2014)  1725--1732

\end{thebibliography}
\end{document}